%% file: sn-article.tex
\documentclass[pdflatex,sn-mathphys-num]{sn-jnl}% Math and Physical Sciences 

\usepackage{graphicx}%
\usepackage{multirow}%
\usepackage{amsmath,amssymb,amsfonts}%
\usepackage{amsthm}%
\usepackage{mathrsfs}%
\usepackage[title]{appendix}%
\usepackage{xcolor}% 
\usepackage{textcomp}%
\usepackage{manyfoot}%
\usepackage{booktabs}%
\usepackage{algorithm}%
\usepackage{algorithmicx}%
\usepackage{algpseudocode}%
\usepackage{listings}%
\usepackage{listings}
\usepackage{color}
\usepackage{xcolor}
\usepackage{graphicx}
\usepackage{float}
\usepackage{physics}
\usepackage{comment}
\usepackage{tcolorbox}
\usepackage{rotating} % for sideways table
\usepackage{bbm} % for the indicator function
\usepackage{upgreek} % for up greek letters

\definecolor{dkblue}{rgb}{0,0.39,0}
\definecolor{gray}{rgb}{0.66,0.66,0.66}
\definecolor{mauve}{rgb}{0.91,0.33,0.50}
\definecolor{gold}{rgb}{1,0.84,0}

\begin{document}

%\title[Proposal: DEVCOM Army Research Laboratory Funding]{DEVCOM Army Research Laboratory Funding -- 
\title{Optimal Path Planning and Cost Minimization for a Drone Delivery System Via Model Predictive Control}
\author*[1]{ \fnm{Muhammad Al-Zafar} \sur{Khan}}\email{Muhammad.Al-ZafarKhan@zu.ac.ae}

\author[1,2]{\fnm{Jamal} \sur{Al-Karaki}}\email{Jamal.Al-Karaki@zu.ac.ae}

\affil*[1]{\orgname{College of Interdisciplinary Studies}, Zayed University, \state{Abu Dhabi}, \country{UAE}}

\affil[2]{\orgname{College of Engineering}, The Hashemite University \state{Zarqa}, \country{Jordan}}

\abstract{In this study, we formulate the drone delivery problem as a control problem and solve it using Model Predictive Control. Two experiments are performed: The first is on a less challenging grid world environment with lower dimensionality, and the second is with a higher dimensionality and added complexity. The MPC method was benchmarked against three popular Multi-Agent Reinforcement Learning (MARL): Independent $Q$-Learning (IQL), Joint Action Learners (JAL), and Value-Decomposition Networks (VDN). It was shown that the MPC method solved the problem quicker and required fewer optimal numbers of drones to achieve a minimized cost and navigate the optimal path.}

\maketitle

\section{Introduction}\label{introduction}
The rapid evolution of e-commerce and the increasing demand for faster, more efficient delivery systems have ushered in a new era in logistics and the shopping experience, which has huge effects on traditional brick-and-mortar shopping centers and malls that have globally reported a decrease in walk-in retail customers since the COVID-19 pandemic \cite{andruetto2023transition}. According to a report by Shopify on their website \cite{shopify_ecommerce_2023}, global e-commerce sales were estimated to be \$6.09 trillion and forecasted to be around \$8.09 trillion in 2028. Within this B2B and B2C value chain, a customer orders via websites or apps and then has it delivered to the address of their choice. This is the process of shipping the package to a centralized depot and then hauling it to them. Traditional delivery methods, reliant on delivery trucks or vans and human drivers, are becoming increasingly unsustainable due to rising costs \cite{ranieri2018review}, environmental concerns \cite{viu2020impact,kiba2021sustainable,gund2024q}, and traffic congestion \cite{anderson2003commerce}. In this context, autonomous drone delivery systems have emerged as a viable solution, promising to revolutionize last-mile logistics by offering faster, cheaper, and more environmentally friendly delivery options. The need for robust and efficient path-planning algorithms becomes paramount as the world moves toward a future where drones replace vans for package delivery \cite{benarbia2021literature}. This paper explores using Model Predictive Control (MPC) to determine optimal delivery paths and minimalistic delivery costs for drones, ensuring safe, efficient, and collision-free operations in complex urban environments.

The shift from traditional delivery methods to drone-based systems is not merely a technological advancement but a fundamental reimagining of commerce that will inevitably occur at some point. Drones offer unparalleled advantages, including reduced delivery times, lower operational costs, and the ability to reach remote or congested areas challenging for ground vehicles. However, the success of drone delivery systems hinges on their ability to navigate dynamic environments, avoid obstacles, and optimize routes in real time. This requires sophisticated control strategies that can handle the complexities of urban airspace, including restricted zones, no-fly areas, and the presence of other drones. 

While Reinforcement Learning (RL), particularly Multi-Agent Reinforcement Learning (MARL), has been widely explored for autonomous systems, they often struggle with scalability, interpretability, and real-time adaptability. RL-based approaches, though powerful and robust, are: 
\begin{enumerate}
\item \textbf{Data Hungry:} MARL algorithms require extensive training data and may fail to generalize to unseen scenarios.
\item \textbf{Non-Stationarity of Environments:} Each agent's policy evolves over time, making the environment non-stationary for every agent. Non-stationarity is unavoidable because other agents are simultaneously learning and changing their policies. From any single agent's perspective, the environment seems to change unpredictably, and the same action in the same state can lead to different outcomes as other agents evolve. 
\item \textbf{Curse of Dimensionality:} The state-action space grows exponentially as more and more agents are added to the system, and therefore joint $Q$-tables and policies become infeasible. 
\item \textbf{Credit Assignment Problem:} If we assume that all agents in the system work together cooperatively, it is very difficult to determine which agent's actions were responsible for particular outcomes. 
\item \textbf{Coordination:} If the agents do not communicate with each other, coordination becomes very challenging. Further, the design of effective communication protocols in order to facilitate coordination is very difficult.
\item \textbf{Exploration vs. Exploitation:} In multiagent settings, balancing exploration (the drones trying new delivery strategies) and exploitation (using known good delivery routes strategies) is very challenging. If not properly balanced, this often leads to convergence of the agents to suboptimal policies. 
\item \textbf{Scalability:} Many of the well-known algorithms work well with small numbers of agents but struggle with scalability.
\item \textbf{Stability and Convergence:} Due to the dynamic nature of the way in which the agent learns, many of the MARL algorithms do not converge and are unstable. 
\end{enumerate}

In contrast, Model Predictive Control (MPC) \cite{richalet1978model,garcia1989model} offers a model-based framework that explicitly accounts for system dynamics, constraints, and objectives, making it inherently more interpretable and reliable for real-time decision-making. MPC's ability to handle constraints -- such as collision avoidance, restricted airspace, and energy efficiency -- makes it particularly well-suited for drone delivery systems. Recently, MPC has seen a resurgence and a particular interest from ML researchers after META's Chief AI Scientist Yann LeCun stated in a Lex Friedman podcast that we abandon RL in favor of MPC and called for \textit{``its use to be minimized because it is incredibly inefficient in terms of samples\ldots''} \cite{lexfriedman2024},  and in lecture notes \cite{LeCun2023lecture}: \textit{``Use RL only when planning doesn’t yield the predicted outcome, to adjust the world model or the critic.''}

The basic premise of MPC is that it uses a dynamic system model to predict future behavior over a finite time horizon. At each control interval, the algorithm solves an optimization problem that minimizes the cost function, which is typically taken to be the tracking error and control effort, while respecting system constraints. Only the first control action from the optimized sequence is applied to the system. The entire process repeats at the next time step with updated measurements, creating a receding horizon approach. Thus, this facilitates MPC handling multivariable systems, anticipating future events, and explicitly accounting for constraints on inputs, states, and outputs.

In this paper, we formulate the problem of a swarm of drones delivering packages from a depot to an environment modeled as a grid world. Within the environment, there are different building types, which have an associated cost of delivery, and the obstacle is restricted airspace, which could arise due to the governmental legislature. We then benchmark the MPC approach against popular MARL algorithms in order to determine the minimum number of drones required to deliver all packages while minimizing the total cost of delivery. 

This paper is divided as follows:

In Sec. \ref{related work}, we present some related studies that used an ML-based approach to accomplish similar tasks to the one we are proposing.

In Sec. \ref{problem formulation}, we mathematically formulate the drone delivery problem as a control problem, given the algorithm that we use to train the model, and we describe the MARL algorithms that we measure against.

In Sec. \ref{experiments}, we apply the algorithm to two environments with varying degrees of difficulty and size.

In Sec. \ref{conclusion}, we provide a closing statement on the research, reflect upon the results obtained, and provide guidelines for future avenues of exploration. 

%=============================================================================
\section{Related Work}\label{related work}
The drone delivery problem, or more generally the \textit{drone routing problem} (DRP), is a neoclassical application of MARL algorithms. For example, in \cite{ding2022combining}, the authors apply a hybrid approach that combines search-based methods with dynamic programming to solve the multiagent pathfinding (MAPF) problem in non-grid spaces. It was shown that this hybrid approach improved the learning capabilities and outperformed existing MAPF methods; a similar feat was achieved in \cite{ding2023marl} when cooperative MARL was applied to solve the DRP. In \cite{kaji2024safe}, the authors tackle the issue of unsafe actions in MARL, namely collisions during navigation, by integrating a safety control method and enhancing the popular QMIX algorithm to Safe QMIX.   

With respect to the applications of MPC to drone delivery problems, the work in \cite{matos2021model} addresses the practical problem of quadcopters running out of battery life midflight. The paper proposes a payload-exchange scheme based on MPC to optimally execute aerial relay maneuvers. There are a plethora of applications to various unmanned aerial vehicle (UAV) applications as evidenced in \cite{richards2004decentralized,nguyen2021model,aliyari2022design,feng2018autonomous}. However, the specific application of MPC to a drone delivery system is not well studied.   

Whereas the techniques of MPC are well-known, the application area is understudied. Thus, the impetus for this research. 

%=============================================================================
\section{The Multi-Agentic Drone Delivery System -- Problem Formulation}\label{problem formulation}

Consider the system $n$ drones that must navigate an environment to deliver packages and return to the warehouse. The environment is composed of unique building types to deliver to, say, homes (with cost $c_{1}$), office buildings (with cost $c_{2}$), shops (with cost $c_{3}$), and so on. In addition, the environment has regions of restricted airspace where the drones cannot fly. The objective is to minimize the number of drones that must be used and the cost of delivery to all building types to find the optimal delivery paths.

We formulate the problem mathematically as an MPC problem. We assume that each drone is modeled as a discrete-time dynamical system. The coordinate positions for drone $i$ at timestamp $k$ are given by
\begin{equation}\label{drone dynamics}
\mathbf{x}_{i}(k+1)=\mathbf{A}^{T}\mathbf{x}_{i}(k)+\mathbf{B}^{T}\mathbf{u}_{i}(k),    
\end{equation}
where $\mathbf{u}_{i}\in\mathbb{R}^{m}$ is the control input, which we take as velocity in this case, and $\mathbf{A}$ and $\mathbf{B}$ are the system transition matrices. 

We consider the environment to be composed of $M$ buildings $\mathcal{B}=\left\{b_{1},b_{2},\ldots,b_{M}\right\}$ with associated costs $\mathcal{C}=\left\{c_{1},c_{2},\ldots,c_{M}\right\}$. We define the restricted airspace by $\mathcal{R}=\bigcup_{k=1}^{K}R_{k}=R_{1}\cup R_{2}\cup\ldots\cup R_{K}$ to be composed of those regions in the delivery region where the drone cannot fly. 

The goal is to solve the optimization problem
\begin{equation}\label{optimization problem}
\underset{\left\{\mathbf{u}_{i}(k),\mathbf{u}_{i}(k+1),\ldots,\mathbf{u}_{i}(k+N-1)\right\}_{i=1}^{n}}{\min}\;\underbrace{\sum_{i=1}^{n}\sum_{j=1}^{M}c_{j}\cdot\mathbbm{1}_{ij}+\overbrace{\lambda\cdot n}^{P}}_{J},    
\end{equation}
where $\mathbbm{1}_{ij}$ is the indicator function that takes on a value of 1 if drone $i$ delivers to building $j$ and is 0 otherwise, the product $\lambda\cdot n$ is the penalty function $P$ that penalizes for the use of extra drones with $0\leq\lambda\leq 1$ is the penalization weight, and $J$ denotes the cost function. 

The optimization problem in \eqref{optimization problem} is solved subject to the constraints
\begin{enumerate}
\item[C1.] \textbf{Drone Dynamics:} The state and control inputs must satisfy the drone dynamics given by \eqref{drone dynamics} $\forall i=1,2,\ldots,n$, with initial condition $\mathbf{x}_{i}(0)=\mathbf{x_{i,0}}$ and final condition $\mathbf{x}_{i}(k+N)=\mathbf{x}_{i,f}$, with $N$ being the horizon lookahead.   
\item[C2.] \textbf{Delivery Costs:} Each building $j$ must be delivered to by exactly one drone,
\begin{equation}
\sum_{i=1}^{n}\mathbbm{1}_{ij}=1,\quad \forall j=1,2,\ldots M.     
\end{equation}
\item[C3.] \textbf{Restricted Airspace:} Drones must avoid restricted airspace regions $\mathcal{R}$. Thus, for the flightpath,
\begin{equation}
\mathbf{x}_{i}(k)\not\in\mathcal{R},\quad \forall i=1,2,\ldots,n,\quad \forall k. 
\end{equation}
\end{enumerate}

\begin{figure}[H]
    \centering
    \includegraphics[width=0.5\linewidth]{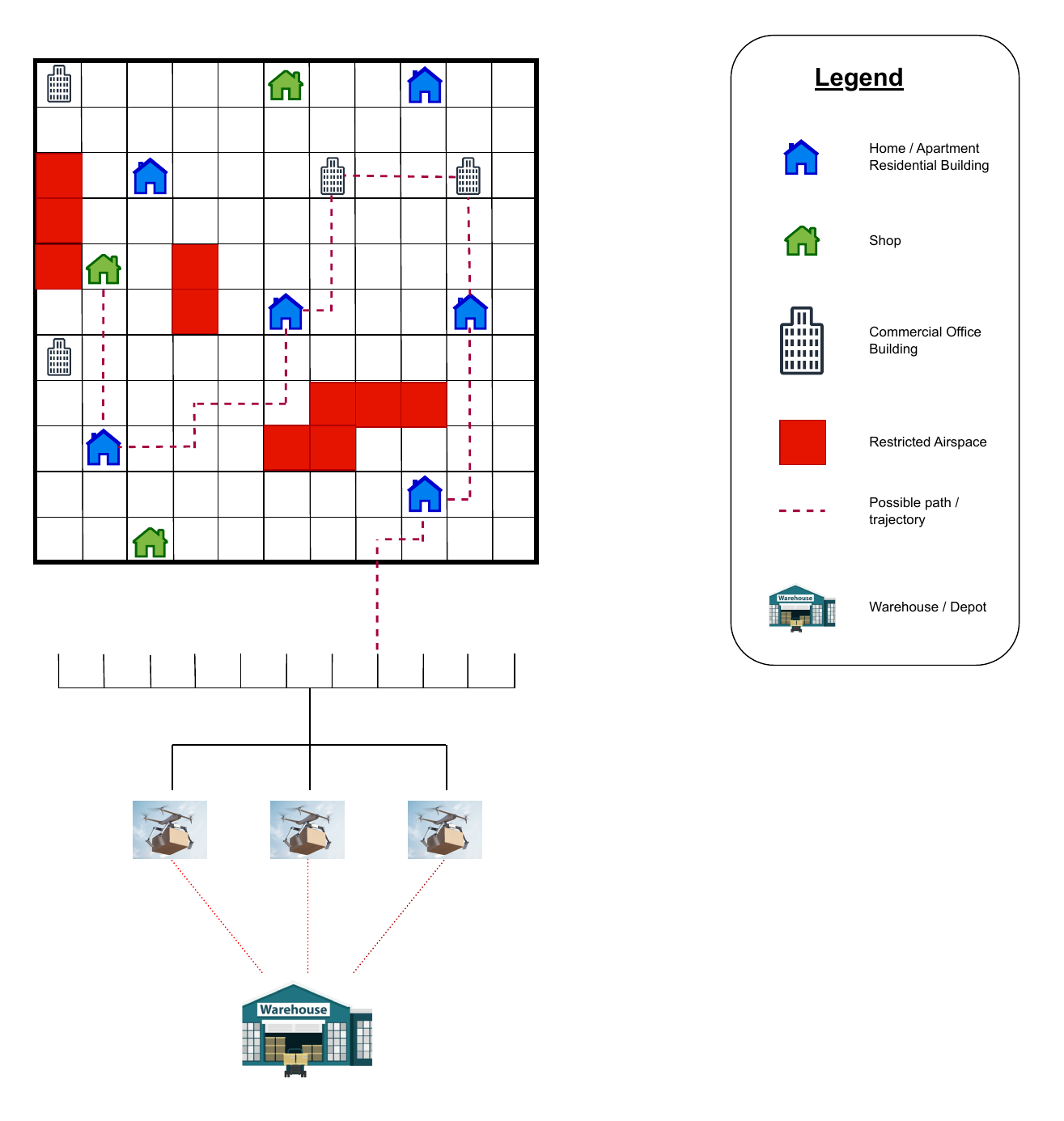}
    \caption{A depiction of a drone delivery system navigating through a grid world environment to deliver packages.}
    \label{fig1}
\end{figure}

We algorithmically encapsulate the learning procedure from MPC for the drone delivery system in Algorithm \ref{algo1} below.

\begin{algorithm}[H]
\caption{Drone Delivery}\label{algo1}
\begin{algorithmic}[1]
\State \textbf{input:} $n$ drones each with position $\mathbf{x}_{i}(t)\in\mathbb{R}^{2}$ and controls $\mathbf{u}_{i}(t)$, $M$ buildings located at $b_{j}\in\mathbb{R}^{2}$ and having cost $c_{j}$, restricted airspace $\mathcal{R}$ with positions $r_{k}\in\mathbb{R}^{2}$, state transition matrices $\mathbf{A},\mathbf{B}\in\mathbb{R}^{2\times 2}$, learning rate $0\leq\alpha\leq1$, lookahead horizon/final timestep $N\in\mathbb{Z}^{+}$, the minimum allowed distance from the restricted airspace $d_{\min}$, penalization parameter $\lambda$
\Repeat
\State calculate delivery cost
\begin{equation*}
J_{\text{delivery}}=\sum_{i=1}^{n}\sum_{j=1}^{M}c_{j}\cdot||\mathbf{x}_{i}(N)-b_{j}||
\end{equation*}
\State calculate restricted airspace cost
\begin{equation*}
J_{\text{restricted airspace}}=\sum_{i=1}^{n}\sum_{t=1}^{N}\sum_{k=1}^{\mathcal{R}}\max\left\{d_{\min}-||\mathbf{x}_{i}(t)-r_{k}||,0\right\}
\end{equation*}
\State calculate penalization cost \Comment{penalizing the control}
\begin{equation*}
J_{\text{penalty}}=\lambda\sum_{i=1}^{n}\sum_{t=1}^{N-1}||\mathbf{u}_{i}(t)||^{2}
\end{equation*}
\State calculate total cost
\begin{equation*}
J=J_{\text{delivery}}+J_{\text{restricted airspace}}+J_{\text{penalty}}    
\end{equation*}
\For {each drone $i=1:n$}
\State learn the new states and new controls 
\begin{equation*}
\begin{aligned}
\mathbf{x}_{i}(t)\gets&\;\mathbf{x}_{i}(t)-\alpha\nabla_{\mathbf{x}_{i}(t)}J, \\
\mathbf{u}_{i}(t)\gets&\;\mathbf{u}_{i}(t)-\alpha\nabla_{\mathbf{u}_{i}(t)}J
\end{aligned}
\end{equation*}
\State update drone system dynamics
\begin{equation*}
\mathbf{x}_{i}(k+1)=\mathbf{A}^{T}\mathbf{x}_{i}(k)+\mathbf{B}^{T}\mathbf{u}_{i}(k)   
\end{equation*}
\EndFor
\Until {$J\to0$ or maximum iterations $N$ is reached}
\end{algorithmic}
\end{algorithm}

In Algorithm \ref{algo1}, given the initial positions of the drones, the buildings, and the restricted airspace, together with the transition matrices, the algorithm calculates the delivery costs by multiplying the cost of delivery to each building type and the distance from where the drone is located to where the building is located and summing all of these up. Similarly, the cost associated with the restricted airspaces is calculated by defining a minimum distance away from the restricted regions and then subtracting the distance of the drone location from the location of the restricted airspace. Finding the maximum value with 0 ensures positivity for all such calculations and summing them up. Thereafter, the control variables are penalized, and the cost of penalizing the controls is calculated. Thus, with the three aforementioned costs, the total cost is calculated. The total cost is then used with gradient-based learning for both the positions and the controls. The system dynamics is then updated according to the standard MPC equation in \eqref{drone dynamics}. The process is repeated until the total cost function is as close to zero as possible or the maximum number of iterations (final timestep) is reached. 

In order to establish a standard for comparing MPC, we benchmark the MPC method against three MARL algorithms. These MARL algorithms are:
\begin{enumerate}
\item \textbf{Independent $Q$-Learning (IQL):} In this algorithm, each agent learns its own action value/$Q$-function independently, ignoring the presence of other agents. For each agent $i\in\mathcal{I}$, the $Q$-value function updated according to
\begin{equation}\label{q-learning}
Q_{i}(s_{i},a_{i})\gets Q_{i}(s_{i},a_{i})+\alpha\left[r_{i}+\gamma\;\underset{a'_{i}}{\max}\;Q_{i}(s'_{i},a'_{i})-Q_{i}(s_{i},a_{i})\right],    
\end{equation}
where $s_{i}$ is the state observed by agent $i$, $a_{i}$ is the action taken by agent $i$, $r_{i}$ is the scalar reward signal received by agent $i$, $s'_{i}$ is the successive/next observed state by agent $i$, $a'_{i}$ is the successive/next action taken by agent $i$, $\alpha$ is the learning rate, and $\gamma$ is the discounting factor. The goal is to minimize the loss of the $i^{\text{th}}$ agent
\begin{equation}\label{q-value loss function}
\mathcal{L}_{i}=\mathbb{E}_{\uptau=\left(s_{i},a_{i},r_{i},s'_{i}\right)}\left\{\left[r_{i}+\gamma\;\underset{a'_{i}}{\max}\;Q_{i}(s'_{i},a'_{i})-Q_{i}(s_{i},a_{i})\right]^{2}\right\}. 
\end{equation}
\item \textbf{Joint Action Learners (JAL):} In this algorithm, agents collectively learn a joint $Q$-function $Q(s,\mathbf{a})$, where $\mathbf{a}=\left(a_{1}, a_{2},\ldots, a_{n}\right)$ is the joint action of all agents. The update rule is analogous to \eqref{q-learning}, expect that the action value function now ingests the joint actions, that is
\begin{equation}
Q(s,\mathbf{a})\gets Q(s,\textbf{a})+\alpha\left[r+\gamma\;\underset{\mathbf{a}'}{\max}\;Q(s',\mathbf{a}')-Q(s,\mathbf{a})\right].  
\end{equation}
Similarly, the goal is to minimize a loss function akin to \eqref{q-value loss function}, except that individual actions of each agent are replaced by the joint action, that is
\begin{equation}\label{JAL loss}
\mathcal{L}=\mathbb{E}_{\uptau=\left(s,\mathbf{a},r,s'\right)}\left\{\left[r+\gamma\;\underset{\mathbf{a}'}{\max}\;Q(s',\mathbf{a}')-Q(s,\mathbf{a})\right]^{2}\right\}.
\end{equation}
\item \textbf{Value-Decomposition Networks (VDN):} In this algorithm, the joint $Q$-value function is decomposed into individual agent $Q$-values. This decomposition is achieved by writing the joint $Q$-value function as the superposition of each agent's $Q$-value function, that is
\begin{equation}\label{superposition of q-functions}
Q(s,\mathbf{a})=\sum_{i=1}^{n}Q_{i}(s_{i},a_{i}).
\end{equation}
The update rule for the $Q$-value function is exactly \eqref{q-learning}, however, the $Q$-term in \eqref{JAL loss} is replaced by the superposition of individual agent's $Q$-values in \eqref{superposition of q-functions}, that is
\begin{equation}
\mathcal{L}=\mathbb{E}_{\uptau=\left(s,\mathbf{a},r,s'\right)}\left\{\left[r+\gamma\;\underset{\mathbf{a}'}{\max}\;\sum_{i=1}^{n}Q_{i}(s'_{i},a'_{i})-\sum_{i=1}^{n}Q_{i}(s_{i},a_{i})\right]^{2}\right\}.
\end{equation}
\end{enumerate}

%============================================================================
\section{Experiments}\label{experiments}
In this section, we perform four experiments each on two environments: The former being a less complicated environment with lower dimensionality and a fewer number of delivery locations and restricted airspace, and the latter being more complicated with an increased number of delivery locations and restricted airspace. The MPC model, trained using Algorithm \ref{algo1}, is benchmarked against the MARL algorithms discussed in Sec. \ref{problem formulation}: IQL, JAL, and VDN. The training results are compared against three metrics: The minimum number of drones required to achieve an optimal solution, the minimum total cost of delivery, and the time it takes to converge to the optimal solution. 

\subsection{Environment 1: Relatively Simple}
\begin{figure}[H]
    \centering
    \includegraphics[width=1.2\linewidth]{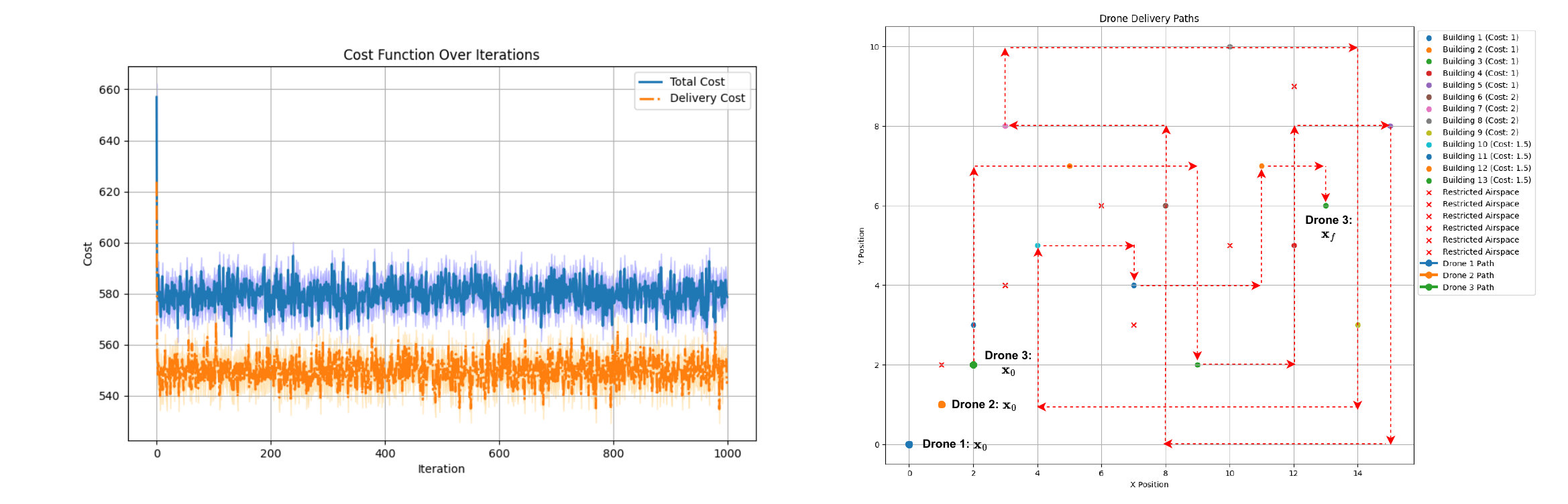}
    \caption{\textbf{Left:} Cost function indicating the total cost and delivery cost. \textbf{Right:} An optimal path that traverses all delivery locations.}
    \label{fig2}
\end{figure}

We consider the grid world environment similar to the one shown in Fig. \ref{fig2}. The environment has three classes of delivery locations to dispatch to:
\begin{enumerate}
\item \textbf{Homes:} Located at $\left\{(2, 3); (5, 7); (9, 2); (12, 5); (15, 8)\right\}$ with a cost of delivery of 1. 
\item \textbf{Offices:} Located at $\left\{(8, 6); (3, 8); (10, 10); (14, 3)\right\}$ with a cost of delivery of 2.
\item \textbf{Shops:} Located at $\left\{(4, 5); (7, 4); (11, 7); (13, 6)\right\}$ with a cost of delivery of 1.5. 
\end{enumerate}

Further, we have restricted airspace located at $\left\{(3, 4); (6, 6); (1, 2); (7, 3); (10, 5); (12, 9)\right\}$ and we add the condition that the drone should maintain a minimum distance of 1 unit away. We arbitrarily set the initial positions of drones 1, 2, and 3 as $\left(0,0\right); \left(1,1\right); \left(2,2\right)$, respectively. 

We initialize the model parameters as
\begin{equation*}
\mathbf{A}=\mathbf{B}=
\begin{pmatrix}
1  &0 \\
0 &1 
\end{pmatrix}, 
\quad N=20, \quad n=3,\quad \lambda=30. 
\end{equation*}
We compare the results from the MPC method with popular methods from MARL in Tab. \ref{tab1} below. We note that drone 3 achieves the minimum costs. 

\begin{table}[h]
\centering
\begin{tabular}{|l|l|l|l|l|}
\hline 
Metrics &MPC &IQL &JAL &VDN \\ 
\hline 
Optimal number of drones &1 &3 &3 &3  \\
Minimum total cost &563.28 &49 &72 &15  \\
Time to convergence &5.84 s &39.80 s &41.04 s &40.25 s \\
\hline 
\end{tabular}
\caption{Comparison of MPC with MARL algorithms for environment 1.}
\label{tab1}
\end{table}

From Fig. \ref{fig2}, we observe that the optimal path predicted by MPC to traverse environment 1 is
\begin{equation*}
\begin{aligned}
&\underbrace{\textcolor{red}{\left(2,2\right)}}_{\text{start}}\to\left(2,3\right)\to\left(5,7\right)\to\left(9,2\right)\to\left(12,5\right)\to\left(15,8\right)\to\left(8,6\right)\to\left(3,8\right) \\
&\to\left(10,10\right)\to\left(14,3\right)\to\left(4,5\right)\to\left(7,4\right)\to\left(11,7\right)\to\underbrace{\textcolor{red}{\left(13,6\right)}}_{\text{end}}. 
\end{aligned}   
\end{equation*}

\begin{figure}[H]
\centering
\includegraphics[width=1.2\linewidth]{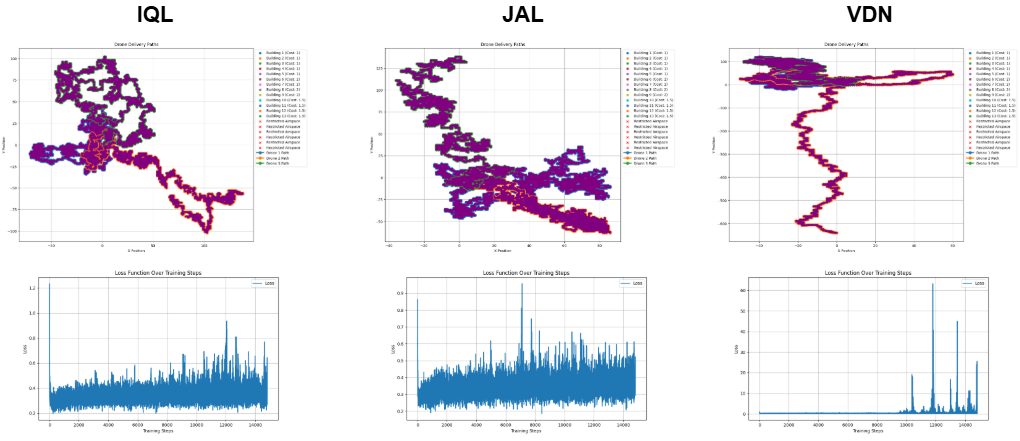}
\caption{Optimal paths and cost functions for the various MARL algorithms applied to environment 1.}
\label{fig3}
\end{figure}

We observe that all three MARL methods require 3 drones each to deliver the packages optimally and cheaply. However, they take significantly longer than the MPC method to converge to the optimal solution. One then needs to balance whether using one or multiple drones for the task is feasible and finding the optimal solution quickly is important. This is a simple task, and it could get much more complicated as the environment becomes more complicated. 

From Fig. \ref{fig3}, we notice from the loss function of the IQL algorithm we observe highly erratic behavior, with the loss function spiking at certain training steps. This is in keeping with the underlying theory that states that the IQL algorithm is easy to implement but suffers from non-stationarity. The JAL algorithm suffers from a similar erratic loss function. As the number of agents in the system increases, the algorithm drastically suffers from the curse of dimensionality. The VDN algorithm displays a relatively regular loss with spikes in the latter part of the training. Based on the loss of all three methods, we can conclude that VDN is the most stable method and can easily be extended to include more agents due to its scalable nature.  

It is important to note that in Fig. \ref{fig2}, the optimal path is sketched as rectilinear lines in the lattice. However, as we can see from the optimal paths in Fig. \ref{fig3}, it need not be rectilinear. In fact, we can confidently say that the geodesic path that minimizes the cost in most scenarios is curvilinear. 

\subsection{Environment 2: More Complex}
In this situation, we modify the environment to add more complexity and increase its dimensionality. Specifically, we increased the number of buildings to be delivered to and increased the amount of restricted airspace there; however, we kept the cost of delivery to each building type the same. The locations of the buildings are now at:
\begin{enumerate}
\item \textbf{Homes:} Located at $\left\{(2, 3); (5, 7); (9, 2); (12, 5); (15, 8); (18, 10); (20, 4); (22, 7); (25, 9)\right\}$. 
\item \textbf{Offices:} Located at $\left\{(8, 6); (3, 8); (10, 10); (14, 3); (17, 5); (19, 8); (21, 2); (24, 6)\right\}$. 
\item \textbf{Shops:} Located at $\left\{(4, 5); (7, 4); (11, 7); (13, 6); (16, 9); (20, 3); (23, 5); (26, 8)\right\}$.
\end{enumerate}

The restricted airspace is now extended to include more regions. These are located at
\begin{equation*}
\begin{aligned}
&\left\{(3, 4); (6, 6); (1, 2); (7, 3); (10, 5); (12, 9); (15, 2); (18, 7); (20, 5); \right. \\
&\left.(22, 3); (24, 8); (26, 4); (28, 6); (30, 3); (32, 7); (34, 5); (36, 9)\right\}.        
\end{aligned}
\end{equation*}

We initialize the model parameters as
\begin{equation*}
\mathbf{A}=\mathbf{B}=
\begin{pmatrix}
1 &0 \\
0 &1
\end{pmatrix}, \quad 
N=30,\quad n=5, \quad\lambda=10.
\end{equation*}
We arbitrarily set the initial positions of drones 1--5 as: $\left\{(0, 0); (1, 1); (2, 2); (3, 3); (4, 4)\right\}$, respectively. 

Performing a similar comparison of MPC with the various MARL techniques as in Tab. \ref{tab1}, we summarize our results in Tab. \ref{tab2} below.

\begin{table}[h]
\centering
\begin{tabular}{|l|l|l|l|l|}
\hline 
Metrics &MPC &IQL &JAL &VDN \\ 
\hline 
Optimal number of drones &2 &5 &5 &5  \\
Minimum total cost &2585.41 &405.50 &433.50 &65.50  \\
Time to convergence &31.73 s &179.69 s &182.94 s &175.23 s \\
\hline 
\end{tabular}
\caption{Comparison of MPC with MARL algorithms for environment 1.}
\label{tab2}
\end{table}

According to the MPC method, the optimal path taken by the two drones is:
\begin{enumerate}
\item \textbf{Drone 3:} $\textcolor{red}{\left(2,2\right)}\to\textcolor{red}{2,3}$.
\item \textbf{Drone 5:} 
\begin{equation*}
\begin{aligned}
&\textcolor{red}{\left(4,4\right)}\to\left(5,7\right)\to\left(9,2\right)\to\left(12,5\right)\to\left(15,8\right)\to\left(18,10\right)\to\left(20,4\right)\to\left(22,7\right)\to\left(15,8\right) \\
&\to\left(25,9\right)\to\left(8,6\right)\to\left(3,8\right)\to\left(10,10\right)\to\left(14,3\right)\to\left(17,5\right)\to\left(19,8\right)\to\left(21,2\right) \\
&\to\left(24,6\right)\to\left(4,5\right)\to\left(7,4\right)\to\left(11,7\right)\to\left(13,6\right)\to\left(16,9\right)\to\left(20,3\right)\to\left(23,5\right)\to\textcolor{red}{\left(26,8\right)}.
\end{aligned}
\end{equation*}
\end{enumerate}

\begin{figure}[H]
\centering
\includegraphics[width=1.1\linewidth]{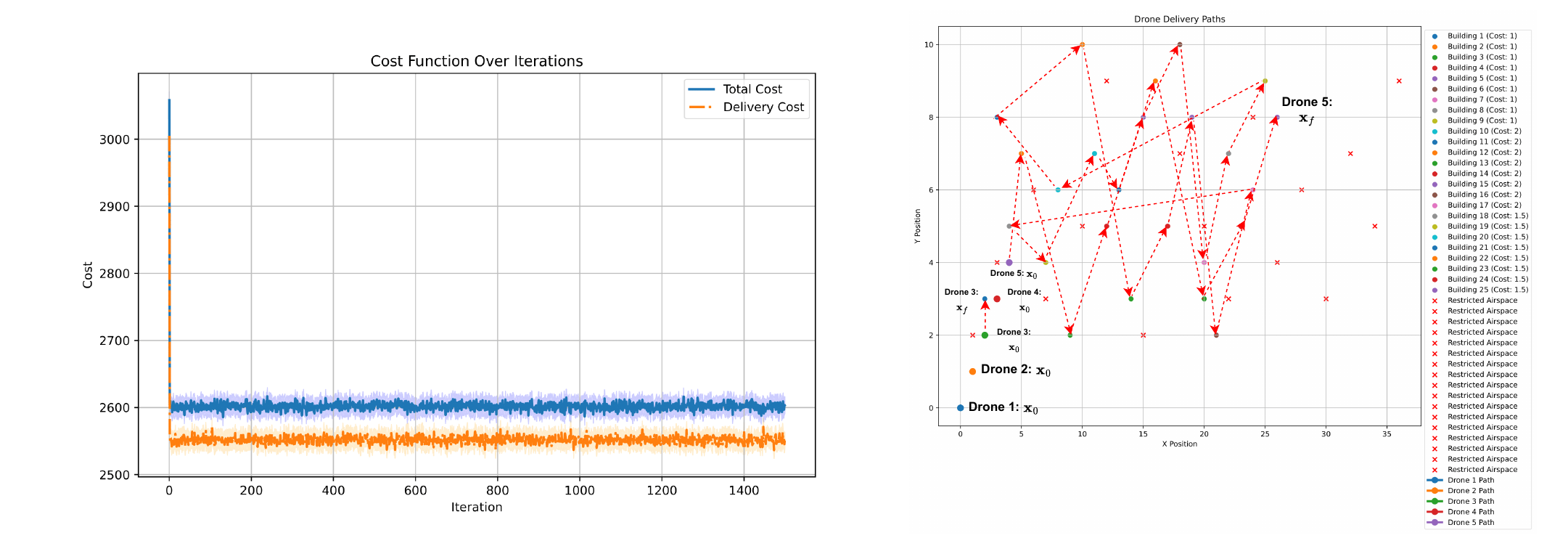}
\caption{\textbf{Left:} Cost function indicating the total cost and delivery cost. \textbf{Right:} A geodesic optimal path traversing all delivery locations.}
\label{fig4}
\end{figure}
From Fig. \ref{fig4}, we see that we optimally only require two drones to minimize the total cost. For the strategically placed initial locations of drone 3 and drone 5, it is possible to sweep out a minimal path. Drone 3 is simply required to execute one maneuver, whilst drone 5 does all the heavy lifting and delivers to all other locations. Additionally, we notice that with the MPC method, the cost function drops from an initial high value to a lower value and stabilizes without erratic spiking. This, in our opinion, is one of the advantages of MPC over MARL.   

\begin{figure}[H]
\centering
\includegraphics[width=1.2\linewidth]{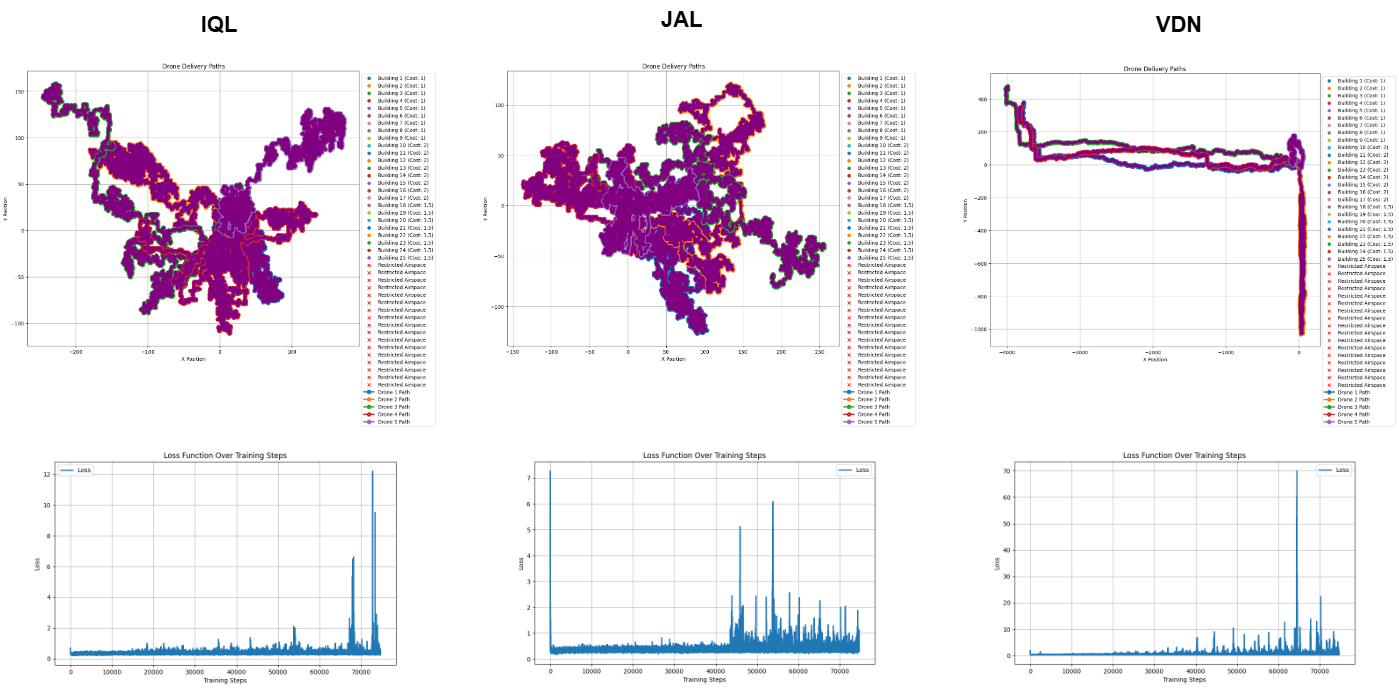}
\caption{Optimal paths and cost functions for various MARL algorithms applied to environment 2.}
\label{fig5}
\end{figure}

From Fig. \ref{fig5}, we observe that the cost functions start low in the IQL and VDN algorithms and then erratically spike towards the latter part of the training. For the JAL algorithm, the cost is initially very high and then drops and maintains a low value. Towards the latter iterations in the training, the cost spikes erratically and stays at high values. In Tab. \ref{tab2}, we see that the MPC algorithm has the highest cost value for training. However, it converges to an optimal solution in the minimum amount of time and requires the least number of drones to achieve optimality. The IQL, JAL, and VDN algorithms take roughly the same time to train. However, the VDN algorithm has the minimum cost and is the quickest of the three. We see that the MARL algorithm requires 5 drones to reach optimality, whereas the MPC method can achieve optimality with just 2 drones, admittedly at a higher cost of delivery.

%============================================================================
\section{Conclusion}\label{conclusion}
We have seen that in both experiments, the MPC method has clearly outperformed the MARL algorithms in terms of how quickly it converged to a solution, requiring fewer drones and producing the optimal path. However, we note that the MARL algorithms were better in terms of producing lower delivery costs. However, we cannot conclude that MARL is better than MPC because all the MARL algorithms required more drones than the MPC method, and required them to work cooperatively to achieve the lower cost. Based on this, we can say that MPC is more efficient. 

In future studies, we would like to extend the MPC methods and create a broader class of MARL benchmarking algorithms, namely QMIX -- which extends VDN by introducing a mixing network that nonlinearly combines individual agent $Q$-values by assuming that all $n$ drones work cooperatively, we hope to use COMA in order to address the credit assignment problem via counterfactual baselines; to model the environment to have a continuous action space and then using MADDPG with a centralized critic that considers the actions of all agents; to introduce inter-drone communication and test an algorithm like RIAL to ascertain the effect of this communication on policy improvement.  

%=============================================================================
% \bibliography{sn-bibliography}
\input sn-article.bbl

%=============================================================================

\end{document}

%% file: sn-article.bbl

%% file: sn-article.bbl
\begin{thebibliography}{20}
\ifx \bisbn   \undefined \def \bisbn  #1{ISBN #1}\fi
\ifx \binits  \undefined \def \binits#1{#1}\fi
\ifx \bauthor  \undefined \def \bauthor#1{#1}\fi
\ifx \batitle  \undefined \def \batitle#1{#1}\fi
\ifx \bjtitle  \undefined \def \bjtitle#1{#1}\fi
\ifx \bvolume  \undefined \def \bvolume#1{\textbf{#1}}\fi
\ifx \byear  \undefined \def \byear#1{#1}\fi
\ifx \bissue  \undefined \def \bissue#1{#1}\fi
\ifx \bfpage  \undefined \def \bfpage#1{#1}\fi
\ifx \blpage  \undefined \def \blpage #1{#1}\fi
\ifx \burl  \undefined \def \burl#1{\textsf{#1}}\fi
\ifx \doiurl  \undefined \def \doiurl#1{\url{https://doi.org/#1}}\fi
\ifx \betal  \undefined \def \betal{\textit{et al.}}\fi
\ifx \binstitute  \undefined \def \binstitute#1{#1}\fi
\ifx \binstitutionaled  \undefined \def \binstitutionaled#1{#1}\fi
\ifx \bctitle  \undefined \def \bctitle#1{#1}\fi
\ifx \beditor  \undefined \def \beditor#1{#1}\fi
\ifx \bpublisher  \undefined \def \bpublisher#1{#1}\fi
\ifx \bbtitle  \undefined \def \bbtitle#1{#1}\fi
\ifx \bedition  \undefined \def \bedition#1{#1}\fi
\ifx \bseriesno  \undefined \def \bseriesno#1{#1}\fi
\ifx \blocation  \undefined \def \blocation#1{#1}\fi
\ifx \bsertitle  \undefined \def \bsertitle#1{#1}\fi
\ifx \bsnm \undefined \def \bsnm#1{#1}\fi
\ifx \bsuffix \undefined \def \bsuffix#1{#1}\fi
\ifx \bparticle \undefined \def \bparticle#1{#1}\fi
\ifx \barticle \undefined \def \barticle#1{#1}\fi
\bibcommenthead
\ifx \bconfdate \undefined \def \bconfdate #1{#1}\fi
\ifx \botherref \undefined \def \botherref #1{#1}\fi
\ifx \url \undefined \def \url#1{\textsf{#1}}\fi
\ifx \bchapter \undefined \def \bchapter#1{#1}\fi
\ifx \bbook \undefined \def \bbook#1{#1}\fi
\ifx \bcomment \undefined \def \bcomment#1{#1}\fi
\ifx \oauthor \undefined \def \oauthor#1{#1}\fi
\ifx \citeauthoryear \undefined \def \citeauthoryear#1{#1}\fi
\ifx \endbibitem  \undefined \def \endbibitem {}\fi
\ifx \bconflocation  \undefined \def \bconflocation#1{#1}\fi
\ifx \arxivurl  \undefined \def \arxivurl#1{\textsf{#1}}\fi
\csname PreBibitemsHook\endcsname

%%% 1
\bibitem[\protect\citeauthoryear{Andruetto et~al.}{2023}]{andruetto2023transition}
\begin{barticle}
\bauthor{\bsnm{Andruetto}, \binits{C.}},
\bauthor{\bsnm{Bin}, \binits{E.}},
\bauthor{\bsnm{Susilo}, \binits{Y.}},
\bauthor{\bsnm{Pernest{\aa}l}, \binits{A.}}:
\batitle{Transition from physical to online shopping alternatives due to the covid-19 pandemic-a case study of italy and sweden}.
\bjtitle{Transportation Research Part A: Policy and Practice}
\bvolume{171},
\bfpage{103644}
(\byear{2023})
\end{barticle}
\endbibitem

%%% 2
\bibitem[\protect\citeauthoryear{Shopify}{2023}]{shopify_ecommerce_2023}
\begin{botherref}
\oauthor{\bsnm{Shopify}}:
Global Ecommerce Sales (2023-2026): Statistics and Trends.
Shopify. Accessed: March 17, 2025.
\url{https://www.shopify.com/blog/global-ecommerce-sales}
Accessed 2025-03-17
\end{botherref}
\endbibitem

%%% 3
\bibitem[\protect\citeauthoryear{Ranieri et~al.}{2018}]{ranieri2018review}
\begin{barticle}
\bauthor{\bsnm{Ranieri}, \binits{L.}},
\bauthor{\bsnm{Digiesi}, \binits{S.}},
\bauthor{\bsnm{Silvestri}, \binits{B.}},
\bauthor{\bsnm{Roccotelli}, \binits{M.}}:
\batitle{A review of last mile logistics innovations in an externalities cost reduction vision}.
\bjtitle{Sustainability}
\bvolume{10}(\bissue{3}),
\bfpage{782}
(\byear{2018})
\end{barticle}
\endbibitem

%%% 4
\bibitem[\protect\citeauthoryear{Viu-Roig and Alvarez-Palau}{2020}]{viu2020impact}
\begin{barticle}
\bauthor{\bsnm{Viu-Roig}, \binits{M.}},
\bauthor{\bsnm{Alvarez-Palau}, \binits{E.J.}}:
\batitle{The impact of e-commerce-related last-mile logistics on cities: A systematic literature review}.
\bjtitle{Sustainability}
\bvolume{12}(\bissue{16}),
\bfpage{6492}
(\byear{2020})
\end{barticle}
\endbibitem

%%% 5
\bibitem[\protect\citeauthoryear{Kiba-Janiak et~al.}{2021}]{kiba2021sustainable}
\begin{barticle}
\bauthor{\bsnm{Kiba-Janiak}, \binits{M.}},
\bauthor{\bsnm{Marcinkowski}, \binits{J.}},
\bauthor{\bsnm{Jagoda}, \binits{A.}},
\bauthor{\bsnm{Skowro{\'n}ska}, \binits{A.}}:
\batitle{Sustainable last mile delivery on e-commerce market in cities from the perspective of various stakeholders. literature review}.
\bjtitle{Sustainable Cities and Society}
\bvolume{71},
\bfpage{102984}
(\byear{2021})
\end{barticle}
\endbibitem

%%% 6
\bibitem[\protect\citeauthoryear{Gund and Daniel}{2024}]{gund2024q}
\begin{barticle}
\bauthor{\bsnm{Gund}, \binits{H.P.}},
\bauthor{\bsnm{Daniel}, \binits{J.}}:
\batitle{Q-commerce or e-commerce? a systematic state of the art on comparative last-mile logistics greenhouse gas emissions literature review}.
\bjtitle{International Journal of Industrial Engineering and Operations Management}
\bvolume{6}(\bissue{3}),
\bfpage{185}--\blpage{207}
(\byear{2024})
\end{barticle}
\endbibitem

%%% 7
\bibitem[\protect\citeauthoryear{Anderson et~al.}{2003}]{anderson2003commerce}
\begin{barticle}
\bauthor{\bsnm{Anderson}, \binits{W.P.}},
\bauthor{\bsnm{Chatterjee}, \binits{L.}},
\bauthor{\bsnm{Lakshmanan}, \binits{T.}}:
\batitle{E-commerce, transportation, and economic geography}.
\bjtitle{Growth and change}
\bvolume{34}(\bissue{4}),
\bfpage{415}--\blpage{432}
(\byear{2003})
\end{barticle}
\endbibitem

%%% 8
\bibitem[\protect\citeauthoryear{Benarbia and Kyamakya}{2021}]{benarbia2021literature}
\begin{barticle}
\bauthor{\bsnm{Benarbia}, \binits{T.}},
\bauthor{\bsnm{Kyamakya}, \binits{K.}}:
\batitle{A literature review of drone-based package delivery logistics systems and their implementation feasibility}.
\bjtitle{Sustainability}
\bvolume{14}(\bissue{1}),
\bfpage{360}
(\byear{2021})
\end{barticle}
\endbibitem

%%% 9
\bibitem[\protect\citeauthoryear{Richalet et~al.}{1978}]{richalet1978model}
\begin{barticle}
\bauthor{\bsnm{Richalet}, \binits{J.}},
\bauthor{\bsnm{Rault}, \binits{A.}},
\bauthor{\bsnm{Testud}, \binits{J.}},
\bauthor{\bsnm{Papon}, \binits{J.}}:
\batitle{Model predictive heuristic control}.
\bjtitle{Automatica (journal of IFAC)}
\bvolume{14}(\bissue{5}),
\bfpage{413}--\blpage{428}
(\byear{1978})
\end{barticle}
\endbibitem

%%% 10
\bibitem[\protect\citeauthoryear{Garcia et~al.}{1989}]{garcia1989model}
\begin{barticle}
\bauthor{\bsnm{Garcia}, \binits{C.E.}},
\bauthor{\bsnm{Prett}, \binits{D.M.}},
\bauthor{\bsnm{Morari}, \binits{M.}}:
\batitle{Model predictive control: Theory and practice—a survey}.
\bjtitle{Automatica}
\bvolume{25}(\bissue{3}),
\bfpage{335}--\blpage{348}
(\byear{1989})
\end{barticle}
\endbibitem

%%% 11
\bibitem[\protect\citeauthoryear{Fridman}{2024}]{lexfriedman2024}
\begin{botherref}
\oauthor{\bsnm{Fridman}, \binits{L.}}:
Yann LeCun: Why RL is overrated.
YouTube video, accessed: 2025-03-17
(2024).
\url{https://www.youtube.com/watch?v=AumK4yYBBnU}
\end{botherref}
\endbibitem

%%% 12
\bibitem[\protect\citeauthoryear{LeCun}{2023}]{LeCun2023lecture}
\begin{botherref}
\oauthor{\bsnm{LeCun}, \binits{Y.}}:
Do large language models need sensory grounding for meaning and understanding?
Accessed: 2025-03-17
(2023).
\url{https://drive.google.com/file/d/1BU5bV3X5w65DwSMapKcsr0ZvrMRU_Nbi/view}
\end{botherref}
\endbibitem

%%% 13
\bibitem[\protect\citeauthoryear{Ding et~al.}{2022}]{ding2022combining}
\begin{bchapter}
\bauthor{\bsnm{Ding}, \binits{S.}},
\bauthor{\bsnm{Aoyama}, \binits{H.}},
\bauthor{\bsnm{Lin}, \binits{D.}}:
\bctitle{Combining multiagent reinforcement learning and search method for drone delivery on a non-grid graph}.
In: \bbtitle{International Conference on Practical Applications of Agents and Multi-Agent Systems},
pp. \bfpage{112}--\blpage{126}
(\byear{2022}).
\bcomment{Springer}
\end{bchapter}
\endbibitem

%%% 14
\bibitem[\protect\citeauthoryear{Ding et~al.}{2023}]{ding2023marl}
\begin{bchapter}
\bauthor{\bsnm{Ding}, \binits{S.}},
\bauthor{\bsnm{Aoyama}, \binits{H.}},
\bauthor{\bsnm{Lin}, \binits{D.}}:
\bctitle{Marl 4 drp: benchmarking cooperative multi-agent reinforcement learning algorithms for drone routing problems}.
In: \bbtitle{Pacific Rim International Conference on Artificial Intelligence},
pp. \bfpage{459}--\blpage{465}
(\byear{2023}).
\bcomment{Springer}
\end{bchapter}
\endbibitem

%%% 15
\bibitem[\protect\citeauthoryear{Kaji et~al.}{2024}]{kaji2024safe}
\begin{bchapter}
\bauthor{\bsnm{Kaji}, \binits{M.}},
\bauthor{\bsnm{Lin}, \binits{D.}},
\bauthor{\bsnm{Uwano}, \binits{F.}}:
\bctitle{Safe multi-agent reinforcement learning for drone routing problems}.
In: \bbtitle{International Conference on Principles and Practice of Multi-Agent Systems},
pp. \bfpage{329}--\blpage{334}
(\byear{2024}).
\bcomment{Springer}
\end{bchapter}
\endbibitem

%%% 16
\bibitem[\protect\citeauthoryear{Matos and Guerreiro}{2021}]{matos2021model}
\begin{bchapter}
\bauthor{\bsnm{Matos}, \binits{F.}},
\bauthor{\bsnm{Guerreiro}, \binits{B.}}:
\bctitle{Model predictive control strategies for parcel relay manoeuvres using drones}.
In: \bbtitle{2021 International Young Engineers Forum (YEF-ECE)},
pp. \bfpage{32}--\blpage{37}
(\byear{2021}).
\bcomment{IEEE}
\end{bchapter}
\endbibitem

%%% 17
\bibitem[\protect\citeauthoryear{Richards and How}{2004}]{richards2004decentralized}
\begin{bchapter}
\bauthor{\bsnm{Richards}, \binits{A.}},
\bauthor{\bsnm{How}, \binits{J.}}:
\bctitle{Decentralized model predictive control of cooperating uavs}.
In: \bbtitle{2004 43rd IEEE Conference on Decision and Control (CDC)(IEEE Cat. No. 04CH37601)},
vol. \bseriesno{4},
pp. \bfpage{4286}--\blpage{4291}
(\byear{2004}).
\bcomment{IEEE}
\end{bchapter}
\endbibitem

%%% 18
\bibitem[\protect\citeauthoryear{Nguyen et~al.}{2021}]{nguyen2021model}
\begin{bchapter}
\bauthor{\bsnm{Nguyen}, \binits{H.}},
\bauthor{\bsnm{Kamel}, \binits{M.}},
\bauthor{\bsnm{Alexis}, \binits{K.}},
\bauthor{\bsnm{Siegwart}, \binits{R.}}:
\bctitle{Model predictive control for micro aerial vehicles: A survey}.
In: \bbtitle{2021 European Control Conference (ECC)},
pp. \bfpage{1556}--\blpage{1563}
(\byear{2021}).
\bcomment{IEEE}
\end{bchapter}
\endbibitem

%%% 19
\bibitem[\protect\citeauthoryear{Aliyari et~al.}{2022}]{aliyari2022design}
\begin{barticle}
\bauthor{\bsnm{Aliyari}, \binits{M.}},
\bauthor{\bsnm{Wong}, \binits{W.-K.}},
\bauthor{\bsnm{Bouteraa}, \binits{Y.}},
\bauthor{\bsnm{Najafinia}, \binits{S.}},
\bauthor{\bsnm{Fekih}, \binits{A.}},
\bauthor{\bsnm{Mobayen}, \binits{S.}}:
\batitle{Design and implementation of a constrained model predictive control approach for unmanned aerial vehicles}.
\bjtitle{IEEE Access}
\bvolume{10},
\bfpage{91750}--\blpage{91762}
(\byear{2022})
\end{barticle}
\endbibitem

%%% 20
\bibitem[\protect\citeauthoryear{Feng et~al.}{2018}]{feng2018autonomous}
\begin{barticle}
\bauthor{\bsnm{Feng}, \binits{Y.}},
\bauthor{\bsnm{Zhang}, \binits{C.}},
\bauthor{\bsnm{Baek}, \binits{S.}},
\bauthor{\bsnm{Rawashdeh}, \binits{S.}},
\bauthor{\bsnm{Mohammadi}, \binits{A.}}:
\batitle{Autonomous landing of a uav on a moving platform using model predictive control}.
\bjtitle{Drones}
\bvolume{2}(\bissue{4}),
\bfpage{34}
(\byear{2018})
\end{barticle}
\endbibitem

\end{thebibliography}
